 \documentclass[final]{cvpr}

\usepackage{times}
\usepackage{epsfig}
\usepackage{graphicx}
\usepackage{amsmath}
\usepackage{amssymb}
\usepackage{algorithm}
\usepackage{style}

\usepackage{algpseudocode}
\DeclareMathOperator*{\argmin}{arg\,min}

\usepackage[pagebackref=true,breaklinks=true,colorlinks,bookmarks=false]{hyperref}

\def\cvprPaperID{10436} 
\def\confYear{CVPR 2021}

\begin{document}

\title{Incremental Learning via Rate Reduction}

\author {Ziyang Wu*\\
Cornell\\
{\tt\small zw287@cornell.edu}
\and
Christina Baek*\\
UC Berkeley\\
{\tt\small ke.baek@berkeley.edu}
\and
Chong You\\
UC Berkeley\\
{\tt\small cyou@berkeley.edu}
\and
Yi Ma\\
UC Berkeley\\
{\tt\small yima@eecs.berkeley.edu}}

\maketitle

\begin{abstract}
Current deep learning architectures suffer from catastrophic forgetting, a failure to retain knowledge of previously learned classes when incrementally trained on new classes. The fundamental roadblock faced by deep learning methods is that deep learning models are optimized as ``black boxes'', making it difficult to properly adjust the model parameters to preserve knowledge about previously seen data. To overcome the problem of catastrophic forgetting, we propose utilizing an alternative ``white box'' architecture derived from the principle of rate reduction, where each layer of the network is explicitly computed without back propagation. Under this paradigm, we demonstrate that, given a pretrained network and new data classes, our approach can provably construct a new network that emulates joint training with all past and new classes. Finally, our experiments show that our proposed learning algorithm observes significantly less decay in classification performance, outperforming state of the art methods on MNIST and CIFAR-10 by a large margin and justifying the use of ``white box'' algorithms for incremental learning even for sufficiently complex image data. 

\end{abstract}

\section{Introduction}

Humans are capable of acquiring new information continuously while retaining previously obtained knowledge. This seemingly natural capability, however, is extremely difficult for deep neural networks (DNNs) to achieve. {\em Incremental learning} (IL), also known as continual learning or life-long learning, thus studies the design of machine learning systems that can assimilate new information without forgetting past knowledge. 

In incremental learning, models go through rounds of training sessions to accumulate knowledge for a particular objective (\eg classification). Specifically, under \textit{class} incremental learning (class-IL), an agent has access to training data from a subset of the classes, known as a \textit{task}, at each training session and is evaluated on all seen classes at inference time.
The overarching goal is to precisely fine-tune a model trained on previously seen tasks to additionally classify new classes of data. However, due to the absence of old data, such models often suffer from {\em catastrophic forgetting} \cite{catastrophic}, which refers to a drastic drop in performance after training incrementally on different tasks.

In the last few years, a flurry of continual learning algorithms have been proposed for DNNs, aiming to alleviate the effect of catastrophic forgetting. These methods can be roughly partitioned into three categories: 1) Regularization-based methods that often involve knowledge distillation \cite{lwf,ewc,oewc,si}, 2) Exemplar-based methods that keep partial copies of data from previously learned tasks \cite{icarl,benchmark,bic}, and 3) Modified architectures that attempt to utilize network components specialized for different tasks \cite{pnn,oewc,ltg}. In practice, these algorithms exhibit varying performance across different datasets and their ability to mitigate catastrophic forgetting is inadequate. Factors including domain shift \cite{saenko2010adapting} across tasks and imbalance of new and past classes \cite{bic} are part of the reason.

The fundamental roadblock in deep continual learning is that DNNs are trained and optimized in a ``black box'' fashion. Each model contains millions of mathematical operations and its complexity prevents humans from following the mapping from data input to prediction. Given our current limited understanding of network parameters, it is difficult, if not impossible, to precisely control the parameters of a pre-trained model such that the decision boundary learned fits to new data without losing its understanding of old data. 

In this work, we take a drastically different approach to incremental learning. We avoid ``black box'' architectures entirely, and instead utilize a recently proposed ``white box'' DNN architecture derived from the principle of rate reduction \cite{redunet}. Termed ReduNet, each layer of this DNN can be explicitly computed in a forward-propagation fashion and each parameter has precise statistical interpretations. The so-constructed network is intrinsically suitable for incremental learning because the second-order statistics of any previously-seen training data is preserved in the network parameters to be leveraged for future tasks.

We propose a new incremental learning algorithm utilizing ReduNet to demonstrate the power and scalability of designing more interpretable networks for continual learning. Specifically, we prove that a ReduNet trained incrementally can be constructed to be equivalent to one obtained by joint training, where all data, both new and old, is assumed to be available at training time. Finally, we observe that ReduNet performs significantly better on MNIST \cite{mnist} and CIFAR-10 \cite{cifar10} in comparison to current continual DNN approaches. 

\section{Related Work}
Since the early success of deep learning in classification tasks such as object recognition, attention has lately shifted to the problem of incremental learning in hopes of designing deep learning systems that are capable of continuously adapting to data from non-stationary and changing distributions.

Incremental learning can refer to different problem settings and most studies focus on three widely accepted scenarios \cite{vandeven}.  Most of the earlier works \cite{lwf,pnn,ewc,oewc} study the \textit{task} incremental (task-IL) setting, where a model, after trained on multiple tasks, must be able to classify on data belonging to all the classes it has seen so far. However, the model is additionally provided a task-ID indicating the task or subset of classes each datapoint belongs to. Models trained under this setting are thus required to distinguish among typically only a small number of classes. Recent works \cite{yu2020drift,dmc} explore the more difficult \textit{class} incremental (class-IL) setting,  where task-ID is withheld at inference time. This setting is considerably more difficult since without the task-ID, each datapoint could potentially belong to any of the classes the model has seen so far. The  other  setting,  known  as  \textit{domain}  incremental learning (domain-IL) differs from the previous two settings in that each task consists of all the classes the model needs to learn. Instead, a task-dependent transformation is applied to the data.  For example, each task could contain the same training data rotated by differing degrees and the model must learn to classify images of all possible rotations without access to the task-ID. 

Deep continual learning literature from the last few  years  can  be  roughly  partitioned  into  three  categories as follows:

\textbf{Regularization-based} methods usually attempt to preserve some part of the network parameters deemed important for previously learned tasks. Knowledge distillation \cite{hinton2015distilling} is a popular technique utilized to preserve knowledge obtained in the past. Learning without Forgetting (LwF) \cite{lwf}, for example, attempts to prevent the model parameters from large drifts during the training of the current task by employing cross-entropy loss regularized by a distillation loss. Alternatively, elastic weight consolidation (EWC) \cite{ewc} attempts to curtail learning on weights based on their importance to previously seen tasks. This is done by imposing a quadratic penalty term that encourages weights to move along directions with low Fisher information. Schwarz \etal \cite{oewc} later proposed  an online variant (oEWC) that reduces the cost of estimating the Fisher information matrix. Similarly, Zenke \etal \cite{si} limits the changes of important parameters in the network by using an easy-to-compute surrogate loss during training.

\textbf{Exemplar-based} methods typically use a memory buffer to store a small set of data from previous tasks in order to alleviate catastrophic forgetting. The data stored is used along with the data from the current task to jointly train the model. Rebuffi \etal \cite{icarl} proposed iCaRL which uses a herding algorithm to decide which samples from each class to store during each training session. This technique is combined with regularization with a distillation loss to further encourage knowledge retention \cite{icarl}. A recent work by Wu \etal \cite{bic} achieved further improvements by correcting the bias towards new classes due to data imbalance, which they empirically show causes degradation in performance for large-scale incremental learning settings. This is accomplished by appending a bias-correction layer at the end of the network. Another increasingly popular approach is to train a generative adversarial network (GAN) \cite{kemker2017fearnet, wu2018memory} on previously seen classes and use the generated synthetic data to facilitate training on future tasks.

\textbf{Architecture-based} methods either involve designing specific components in the architecture to retain knowledge of previously seen data or appending new parameters or entire networks when encountering new classes of data. Progressive Neural Network (PNN) \cite{pnn}, for example, instantiates a new network for each task with lateral connection between networks in order to overcome forgetting. This results in the number of networks to grow linearly with respect to the number of tasks as training progresses. Progress \& Compress (P \& C) \cite{oewc} utilizes one network component to learn the new task, then distills knowledge by EWC \cite{ewc} into the main component that aggregates knowledge from previously encountered data. Li \etal \cite{ltg}, proposes a neural architecture search method that utilizes a separate network that learns whether to reuse, adapt, or add certain building blocks of the main classification network for each task encountered.

Our work studies the more difficult class-IL scenario and does not involve regularization or storing any exemplars. Our method thus can be characterized as an architecture-based approach. However, our method differs with the aforementioned works in several important aspects.
First, we use a ``white box'' architecture that is computed exactly in a feed-forward manner. Moreover, the network, when trained under class-IL scenario, can be shown to perform equivalently to one obtained from joint training while most existing works \cite{ltg,oewc,pnn} based on modified architectures target the less challenging task-IL setting.
We discuss the differences in more detail later in Section \ref{reduinc}, after we have introduced our method properly.

\section{Preliminaries}

In this section, we provide a brief background on the principle of rate reduction and the ``white box'' network architecture (i.e., ReduNet) derived from such a principle. 

\subsection{Principle of Rate Reduction}

Given a set of training data $\{\mb x_i\}$ and their corresponding labels $\{\mb y_i\}$, classical deep learning aims to learn a nonlinear mapping $h(\cdot): \mb x \to \mb y$, implemented as a series of simple linear and nonlinear maps, that minimizes the cross-entropy loss. 
One popular way to interpret the role of multiple layers is to consider the output of each intermediate layer as a latent representation space. 
Then, the beginning layers aim to learn a latent representation $\mb z = f(\mb x, \theta) \in \mathbb{R}^d$ that best facilitates the later layers $\mb y = g(\mb z, \mb w)$ for the downstream classification task. 
As a  concrete example, in image recognition tasks, $f(\cdot)$ is a convolutional backbone that encodes an image $\mb x \in \mathbb{R}^{H\times W \times C}$ into a vector representation $\mb z = f(\mb x, \theta) \in \mathbb{R}^d$ and $g(\mb z) = \mb w \cdot \mb z$ is a linear classifier where $\mb w \in \mathbb{R}^{k \times d}$ and $k$ is the number of labels.
Therefore, it is unclear to what extent the feature representation captures any intrinsic structure of the data. 
Recent work \cite{collapse} shows that this direct label fitting leads to a phenomena called \textit{neural collapse}, where within-class variability and structural information are completely suppressed.

To address the aforementioned problem, a recent work by Yu \etal~\cite{mcr} presented a framework for learning useful and geometrically meaningful representation by maximizing the coding rate reduction (i.e., MCR$^2$). 
Given $m$ training samples of $d$ dimension $\mb{X} \in \R^{d \times m}$ that belong to $k$ classes, let $\mb Z = [f(\mb{x}_1,\theta),...,f(\mb{x}_m, \theta)] \in \R^{d \times m}$ be the latent representation. 
Let $\mb \Pi = \{\mb \Pi^j\}_{j=1}^k$ be the membership of the data in the $k$ classes, where each $\mb {\Pi}^j \in \mathbb{R}^{m \times m}$ is a diagonal matrix such that $\mb {\Pi}^j(i,i)$ is the probability of $\mb{x}^i$ belonging to class $j$. 
Given any prescribed quantization error $\epsilon > 0$, let $R(\mb Z, \epsilon)$ be the lossy coding rate function. 
Then $\text{MCR}^2$ aims to learn a feature representation $\mb Z$ by maximizing the following \emph{rate reduction}:
\begin{align}\label{eq:mcr}
    \Delta  R(\mb Z) &=  R(\mb Z) -  R_{c}(\mb Z, \mb \Pi),
\end{align}
where
\begin{align}
     R(\mb Z) &= \frac{1}{2}\log\det\left(\mb I + \alpha \mb Z \mb Z^\top\right), \quad \text{and} \\
     R_{c}(\mb Z, \mb \Pi) &= \sum_{j=1}^k \frac{\gamma_j}{2}\log\det\left(\mb I + \alpha_j \mb Z \mb \Pi^j \mb Z^\top\right).
\end{align}
In above, we denote $\alpha = d/(m\epsilon^2)$, $\alpha_j = d/(\text{tr}(\mb \Pi^j)\epsilon^2)$, and $\gamma_j = \text{tr}(\mb \Pi^j) /m$\footnote {Before computing $\Delta R$, $\mb Z$ must first be normalized either by projecting each feature $\mb{z}_i$ onto the unit sphere $\mathbb{S}^{d-1}$ or by imposing the Frobenius norm of class features $\mb Z^j = \mb Z \mb \Pi^j$ to scale with the number of samples in class $j$: $\|{\mb Z^j}\|_{F}^{2} = m_{j} = \text{tr}(\mb \Pi^j)$.}. $R(\mb Z)$, known as the \textit{expansion} term, represents the total coding length of all features $\mb Z$ while $R_c(\mb Z, \mb \Pi)$, named \textit{compression} term, measures the sum of coding lengths of each latent class distribution. They are called expansion and compression terms respectively, since to maximize $\Delta R$, the first coding rate term is maximized and the second coding rate term is minimized. This coding rate measure utilizes local $\epsilon$-ball packing to estimate the coding rate of the latent distribution from finite samples. 

In \cite{mcr}, it is demonstrated empirically and theoretically that maximizing $\Delta R(\mb Z)$  enforces the latent class distributions to be low-dimensional subspace-like distributions of approximately $d/k$ dimension. In addition, these class distributions are orthogonal to each other. By doing so, the representation is between-class discriminative, whilst maintaining intra-class diversity. Moreover, these features have precise statistical and geometric interpretations. 

\subsection{Rate Reduction Network} \label{redunet}

While an existing neural network architecture (such as ResNet) can be used for feature learning with MCR$^2$, a follow-up work \cite{redunet} showed that a novel architecture can be explicitly constructed via emulating the projected gradient ascent scheme for maximizing $\Delta R(\mb Z)$.
This produces a ``white box'' network, called ReduNet, which has precise optimization, statistical, and geometric interpretations. 
We review the construction of ReduNet as follows. 


Let $\mb Z$ be initialized as the training data, i.e., $\mb Z_0 = \mb X$.
Then, the projected gradient ascend step for optimizing the rate reduction $\Delta R(\mb Z)$ in \eqref{eq:mcr} is given by
\begin{align} \label{gradientstep}
    \begin{split}
     \mb{Z}_{\ell+1} &\propto \mb{Z}_{\ell} + \eta \left(\frac{\partial \Delta R}{\partial \mb Z}\Bigr|_{\mb{Z}_{\ell}} \right) \\
      &= \mb{Z}_{\ell} + \eta \Big(\mb E_{\ell} \mb Z_{\ell} - \sum_{j=1}^{k} \gamma_j \mb C_{\ell}^j \mb Z_{\ell}^j \Big) \\
     \quad \text{s.t.} \quad &\|{\mb Z_{\ell+1}^j}\|_{F}^{2} = \text{tr}(\mb \Pi^j) = m_{j} \quad \forall j \in \{1,..,k\},
\end{split}
\end{align}
where we use $\mb Z_{\ell}^j = \mb Z_{\ell} \mb {\Pi}^j \in \mathbb{R}^{d \times m}$ to denote the feature matrix associated with the $j$-th class at the $\ell$-th iteration, and $\eta > 0$ is the learning rate. 
The matrices $\mb E_\ell$ and $\mb C_\ell^j$ are obtained by evaluating the derivative $\frac{\partial \Delta R}{\partial \mb Z}$ at $\mb Z_\ell$, given by
\begin{align}
    \mb E_\ell &= \alpha\left(\mb I + \alpha \mb Z_\ell \mb Z_\ell^\top\right)^{-1}, \\
    \mb C_\ell^j &= \alpha_j\left(\mb I + \alpha_j \mb Z_\ell^j \mb{Z}_\ell^{j\top}\right)^{-1}. 
\end{align}


Observe that $\mb E_\ell \in \mathbb{R}^{d\times d}$ is applied to all features $\mb Z_\ell$ and it expands the coding length of the entire data. 
Meanwhile, $\mb C_\ell^j \in \mathbb{R}^{d\times d}$ is applied to features from class $j$, i.e., $\mb Z_\ell^j$, and it compresses the sum of coding lengths of individual classes. 

Once the projected gradient ascent is completed, each gradient step can be interpreted as one layer of a neural network, composed of matrix multiplication and subtraction operators, with $\mb E_\ell$ and $\mb C_\ell^j$ being parameters associated with the $\ell$-th layer.
Then, given a test data $\mb x \in \R^d$, its feature can be computed by setting $\mb z_0 = \mb x$ and iteratively carrying out the following incremental transform
\begin{equation}\label{eq:increment}
    \mb z_{\ell+1} \propto \mb z_{\ell} + \eta \left(\mb E_\ell \mb z_\ell - \sum_{j=1}^k \gamma_j \mb C_\ell^j \mb z_\ell \mb \pi^j(\mb z_\ell)\right).
\end{equation}
Notice that the increment depends on $\mb \pi^j(\mb z_\ell)$, the membership of the feature $\mb z_\ell$, which is unknown for the test data. 
Therefore, \cite{redunet} presented a method that replaces $\mb \pi^j(\mb z_\ell)$ in \eqref{eq:increment} by the following estimated membership
\begin{equation}\label{eq:estimated-membership}
{\hat{\mb \pi}_{\ell}^j}(\mb z) = \frac{\exp{\left({-\lambda k \|\mb C_{\ell}^j \mb z\|}\right)}}{\sum_{j=1}^k \exp{\left(-\lambda k \|\mb C_{\ell}^j \mb z\| \right)}} \in [0,1],
\end{equation}
where $\lambda > 0$ is a confidence parameter.
This leads to a nonlinear operator $\sigma\left( \mb C_\ell^1 \mb z_\ell, \ldots, \mb C_\ell^k \mb z_\ell \right) \doteq \sum_{j=1}^k \gamma_j \mb C_\ell^j \mb z_\ell {\hat{\mb \pi}_{\ell}^j}$ that, after being plugged into \eqref{eq:increment}, produces a nonlinear layer as summarized in Figure~\ref{fig:algorithm}.
Stacking multiple such layers produces a multi-layer neural network for extracting discriminative features.
Then, a nearest subspace classifier as the one presented in Section \ref{nearsub} can classify the data. In addition, each layer is interpretable and computed explicitly.


\begin{figure}[tb]
    \begin{center}
    \includegraphics[scale=0.45]{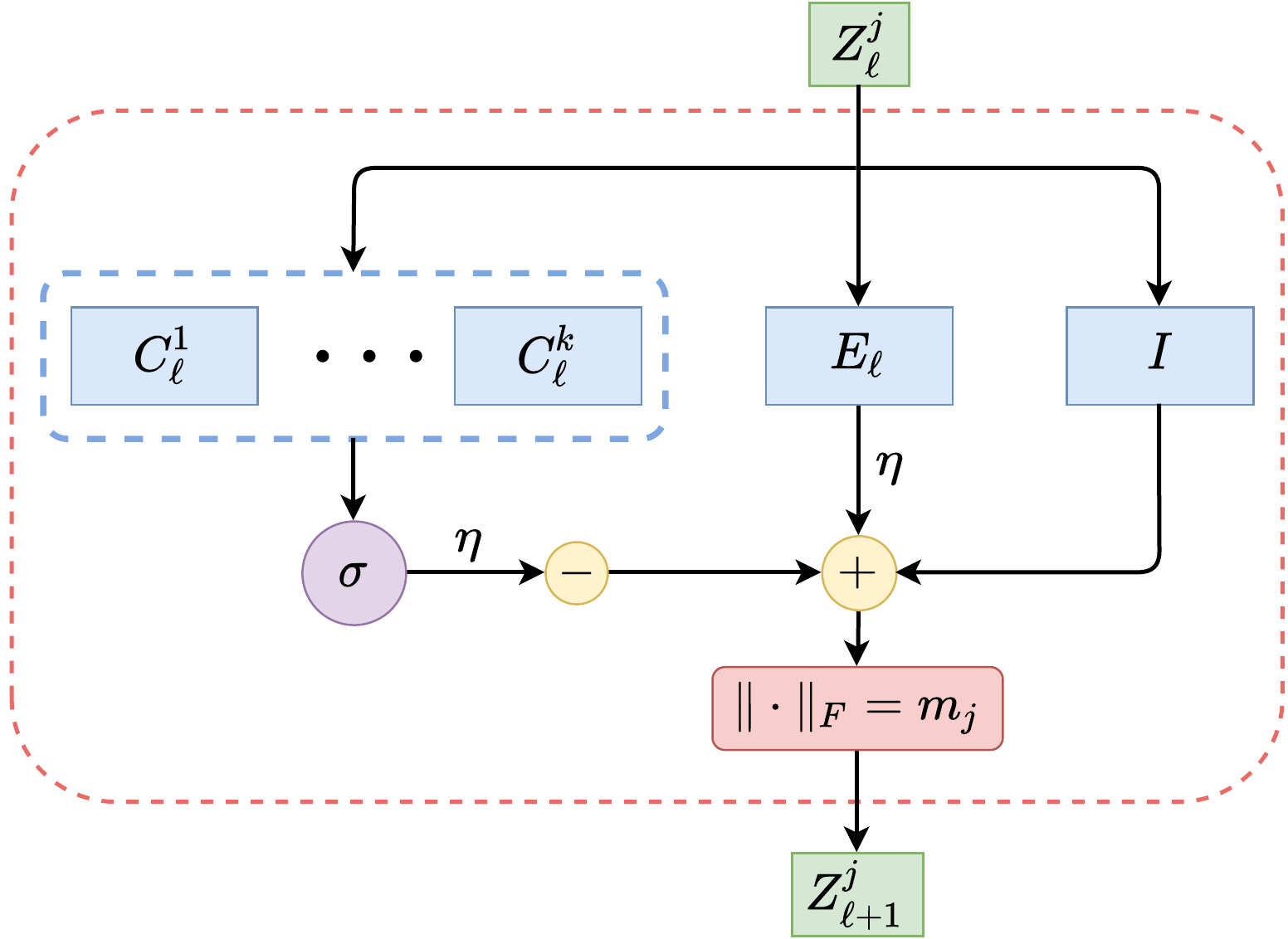}
    \end{center}
    \caption{ReduNet Architecture in which we here adopt a slightly different normalization than \cite{redunet} but is more suitable for the incremental learning as we will see in our derivation.}
    \label{fig:algorithm}
\end{figure}




\section{Incremental Learning with ReduNet}\label{reduinc}
In this paper, we tackle the task of \textit{class} incremental learning, formalized as follows. 
Suppose we have a stream of tasks $\mb{D}^1, \mb{D}^2, \ldots, \mb{D}^t, \ldots$, where each task $\mb{D}^t$ consists of data from $k_t$ classes, i.e, $\mb{D}^{t} = \{\mb{X}^{(t-1)\cdot k_t + 1},\ldots,\mb{X}^{t\cdot k_t }\}$ where $\mb{X}^j$ is a set of points in class $j$. 
The classes in different tasks are assumed to be mutually exclusive. 
Furthermore, it is assumed that the tasks arrive in a online setting, meaning that at timestep $t$ when data $\mb{D}^{t}$ arrives, the data associated with old tasks $\{\mb{D}_i, i < t\}$ becomes unavailable. 
Therefore, the objective is to design a learning system that can adapt the model from the old tasks so as to correctly classify on all tasks hitherto, i.e., $\mb{D}^1,\ldots,\mb{D}^t$. 
In addition, we assume that we are not given the information on the task a test data belongs to, making this problem significantly more challenging than task-IL. 




In this section, we show that ReduNet can perfectly adapt to a new task without forgetting old tasks. Specifically, we present an algorithm to adapt the ReduNet constructed from data $\{\mb{D}_i, i < t\}$ by using only the data in $\mb{D}^{t}$, so that the updated ReduNet is \emph{exactly the same} as the ReduNet constructed as if data from all tasks $\{\mb{D}_i, i \le t\}$ were available. 



\subsection{Derivation of Incrementally-Trained ReduNet}
Without loss of generality, we consider the simple case with two tasks $t$ and $t'$ where $t$ is treated as the old task and $t'$ is treated as the new task. 
Assume that $t$ and $t'$ contain $m_t$, $m_{t'}$ training samples and $k_t$, $k_{t'}$ distinct classes, respectively. 
We denote such training data by $\mb Z_{0,t} \in \R^{d \times m_t}$ (for task $t$) and $\mb Z_{0,t'}\in \R^{d \times m_{t'}}$ (for task $t'$), and assume that they have been normalized by Frobenius norm as described in \eqref{gradientstep}.
For ease of notation, we label the classes as $\{1,...,k_t\}$ for task $t$ and $\{k_t+1,...,k_{t'}+k_t\}$ for task $t'$.

Let $\mb \Theta_t$ be the ReduNet of depth $L$ trained on task $t$ as described in Section  \ref{redunet}. 
Given the new task $\mb Z_{0,t'}$, our objective is to train a network $\mb \Theta_{t \rightarrow {t'}}$ that adapts $\mb \Theta_t$ to have good performance for both tasks $t$ and $t'$.
Next, we show that a network $\mb \Theta_{t \rightarrow {t'}}$ can be constructed from $\mb \Theta_t$ and $\mb Z_{0,t'}$ such that it is \textit{equivalent} to $\mb \Theta$ obtained from training on $\mb Z_0 = [\mb{Z}_{0,t} | \mb{Z}_{0,t'}] \in \R^{d\times m}$ where $m = m_t + m_{t'}$.


To start, consider the initial expansion term $\mb{E}_0 \in \mathbb{R}^{d\times d}$ and compression terms $\mb{C}_{0}^j$ at layer $0$ of the joint network $\mb \Theta$ given by
\begin{align}\label{E_l}
    \begin{split}
        \mb{E}_0 &= \alpha\left(\mb{I}+\alpha\mb{Z}_0\mb{Z}_0^\top\right)^{-1} \\
        &= \alpha\left(\mb{I}+\alpha\big(\mb{Z}_{0,t}\mb{Z}_{0,t}^\top+\mb{Z}_{0,t'}\mb{Z}_{0,t'}^\top\big)\right)^{-1},
    \end{split}
\end{align}
and
\begin{align}\label{C_lj}
    \mb{C}_0^j =
    \begin{cases}
    \alpha_j\left(\mb{I}+\alpha_j\mb{Z}_{0,t}^j \mb{Z}_{0,t}^{j\top}\right)^{-1}, & \text{if}\ j \leq k_t, \\
    \alpha_j\left(\mb{I}+\alpha_j\mb{Z}_{0,t'}^j \mb{Z}_{0,t'}^{j\top}\right)^{-1}, & \text{else},
    \end{cases} 
\end{align}
where $\alpha=d/(m \epsilon^2)$ and $\alpha_j = d/({\text{tr}\mb{(\Pi}^j) \epsilon^2})$. 

Note that the term $\mb{Z}_{0,t'}^j \mb{Z}_{0,t'}^{j\top}$ can be directly computed from input data $\mb{Z}_{0,t'}^{j}$.
On the other hand, the term $\mb Z_{0,t}^j \mb Z_{0,t}^{j\top}$ cannot be directly computed from input data as $\mb{Z}_{0,t}^{j}$ is from the old task, which is no longer available under the IL setup. 
Our key observation is that $\mb Z_{0,t}^j \mb Z_{0,t}^{j\top}$ can be computed from the network $\mb \Theta_t$.
Specifically, by denoting the compression matrices of $\mb \Theta_t$ as $\{\mb C_{\ell,t}^j\}$ for $\ell \in \{0,...,L-1\}$, we have 
\begin{equation}\label{eq:covariance-from-network}
     \mb Z_{0,t}^j \mb Z_{0,t}^{j\top} =  \left((\mb C_{0,t}^{j}/\alpha_j)^{-1} - \mb I\right)/\alpha_j.
\end{equation} 


Next, we show by induction that one can recursively compute $\mb E_\ell$ and $\{\mb C_\ell^j\}$ of $\mb \Theta$ for $\ell > 0$ from \eqref{E_l} and \eqref{C_lj}. 
To construct layer 1 of $\mb \Theta$, we observe that the output features of class $j$ at layer 0 \textit{before} normalization is as follows.
\begin{align}
    \mb P_{0}^j &= (\mb{Z}_{0} + \eta \mb E_{0} \mb Z_{0} - \eta \sum_{i=1}^{k} \gamma_i \mb C_{0}^i \mb Z_{0}^i)\mb \Pi^j \\
    &= \mb{Z}_{0}^j + \eta \mb E_{0} \mb Z_{0}^j - \eta \gamma_j \mb C_{0}^j \mb Z_{0}^j \\
    &= \big(\underbrace{\mb I + \eta \mb{E}_{0} - \eta \gamma_{j}\mb{C}_{0}^j}_{\mb L_0^j \in \mathbb{R}^{d \times d}}\big)\mb{Z}_{0}^j.
\end{align}
Notice the term $\mb L_0^j$ only depends on quantities already obtained at layer 0. To compute $\mb E_1$ and $\mb C_1^j$, we need the covariance matrix of $\mb P_{0}^j$, which we observe to be
\begin{align}
    \mb{T}_{1}^j = \mb P_{0}^j \mb P_{0}^{j\top}
    = \mb L_0^j \mb Z_{0}^j \mb Z_{0}^{j\top} \mb L_0^{j\top}.
\end{align}
Notice that $\mb{T}_{1}^j$ can be expressed with known quantities of $\mb L_0^j$ and $\mb Z_{0,t}^j \mb Z_{0,t}^{j\top}$ if $j \leq k_t$ or $\mb Z_{0,t'}^j \mb Z_{0,t'}^{j\top}$ if $j > k_t$.
The remaining step would be to re-scale $\mb{T}_{1}^j$ as the updated representation $\mb P_{0}^j$ needs to be normalized to get $ \mb Z_{1}^j$.
Recall that we adopt the normalization scheme that imposes the Frobenius norm of each class $\mb Z_j$ to scale with $m_j$:
\begin{align}
    \|{\mb{Z}_{1}^j}\|_F^2 = m_j \iff \text{tr}\big({\mb{Z}_{1}^j}{\mb{Z}_{1}^j}^\top\big) = m_j.
\end{align}
The re-scaling factor is then easy to calculate:
\begin{align}
    \mb Z_{1}^j \mb Z_{1}^{j\top} =  \frac{m_j}{\text{tr}(\mb{T}_{1}^j)}\mb{T}_{1}^j.
\end{align}
From above, we see that we can obtain the correct value of the covariance matrix $\mb Z_{1}^j \mb Z_{1}^{j\top}$, from which we can derive $\mb E_{1}$ and $\mb C_1^j$ for layer 1 of the joint network $\mb \Theta$ and obtain $\mb Z_{2,t'}^j$. 
With these values, we can compute $\mb T_{2}^j$.

By the same logic, we can recursively update $\mb E_\ell$ and $\mb C_\ell^j$ for all $\ell>1$.
Specifically, once we have obtained $\mb Z_{\ell-1}^j \mb Z_{\ell-1}^{j\top}$ and $\mb L_{\ell-1}^j$, it is straightforward to compute  $\mb{T}_{\ell}^j= \mb L_{\ell-1} \mb Z_{\ell-1}^j \mb Z_{\ell-1}^{j\top} \mb L_{\ell-1}^{j\top}$ and therefore obtain
\begin{align}
    \mb Z_{\ell}^j \mb Z_{\ell}^{j\top} =  \frac{m_j}{\text{tr}(\mb{T}_{\ell}^j)}\mb{T}_{\ell}^j.
\end{align}
Note that we never need to access $\mb{Z}_{0,t} \in \mathbb{R}^{d\times m_t}$ directly. 
Instead, we iteratively update the covariance matrix $\mb Z_{\ell-1,t}^j \mb Z_{\ell-1,t}^{j\top} \in \mathbb{R}^{d\times d}$ for each class $j$ using the procedure described.  This concludes our induction and Algorithm \ref{algo} describes the entire training process for incremental learning on two tasks. The procedure is illustrated in Figure \ref{fig:algorithm}.
This procedure can be naturally extended to settings with more than two tasks.

\begin{figure}[tb]
    \begin{center}
    \includegraphics[scale=0.5]{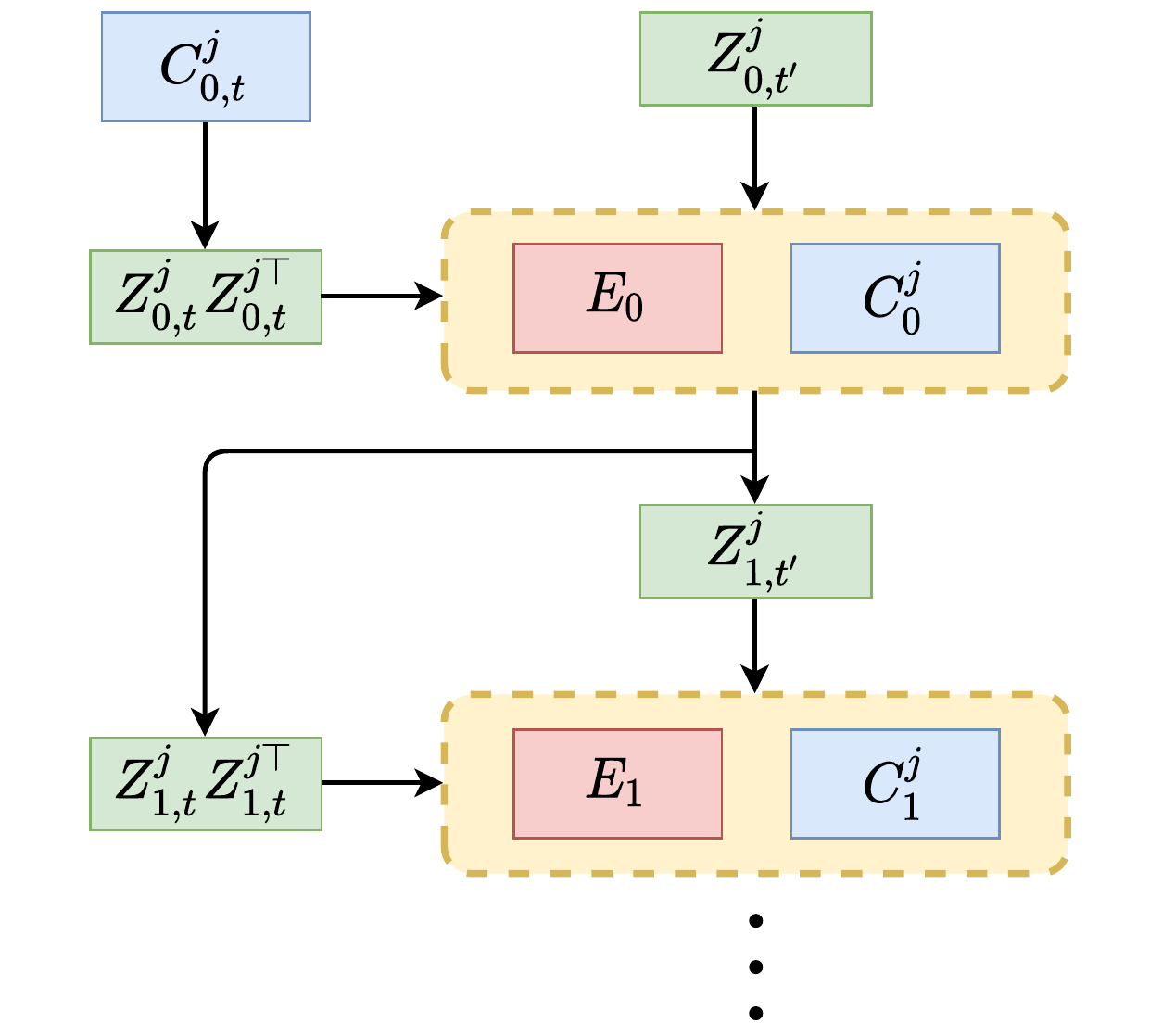}
    \end{center}
    \caption{The joint network can be derived using simply $\mb Z_{0,t} \mb Z_{0,t}^\top$. We do not need the task $t$ data $\mb Z_{0,t}$ directly.}
    \label{fig:algorithm}
\end{figure}
\begin{algorithm}
    \caption{Incremental Learning with ReduNet}
    \begin{algorithmic}[1]
    \State \textbf{Input:} Network $\mb \Theta_t$ with parameters $\mb E_{\ell, t}$ and $\{\mb C_{\ell, t}^j\}$, data $\mb Z_{0,t'}^j$ $\forall j \in \{k_t+1,...,k_t+k_{t'}\}$
    \State Compute $\mb \Sigma_{0,t}^j =\mb{Z}_{0,t}^j \mb{Z}_{0,t}^{j\top}$,    $\forall j \in \{1,...,k_t\}$ by \eqref{eq:covariance-from-network}
    
    \For{$\ell=0,1,2,...,L-1$}
        \State $\mb \Sigma_{\ell,t} = \sum_{j=1}^{k_t} \mb \Sigma_{\ell,t}^{j}$
        \State $\mb \Sigma_{\ell,t'} = \sum_{j=k_t+1}^{k_t + k_{t'}}\mb Z_{\ell,t'}^j\mb Z_{\ell,t'}^{j\top}$
        \State $\mb{E}_{\ell} =  \alpha\big(\mb{I}+\alpha(\mb \Sigma_{\ell,t}+\mb \Sigma_{\ell,t'})\big)^{-1}$
        \State $\mb{L}_{\ell}^j = \mb I + \eta \mb{E}_{\ell} - \eta \gamma_{j}\mb{C}_{\ell}^j \quad \forall j \in \{1,...,k_t\}$
        \For{$j=1,2,...,k_t$}
        \State $\mb{C}_{\ell}^j =  \alpha_j\big(\mb{I}+\alpha_j\mb \Sigma_{\ell}^j\big)^{-1}$
        \State $\mb{T}_{\ell+1,t}^{j}= \mb{L}_{\ell}^j \mb \Sigma_{\ell,t}^j \mb{L}_{\ell}^{j\top}$
        \State $\mb \Sigma_{\ell+1,t}^{j} = \frac{m_j}{\text{tr}\left(\mb{T}_{\ell+1,t}^j\right)}\mb{T}_{\ell+1,t}^j$
     \EndFor
    
    \State $\mb{Z}_{\ell+1,t'} \propto \mb{Z}_{\ell,t'} + \eta \mb E_{\ell} \mb Z_{\ell,t'} - \eta \sum\limits_{i=k_t+1}^{k_t+k_{t'}} \gamma_i \mb C_{\ell}^i \mb Z_{\ell,t'}^i$
    \State$\quad \quad \quad \quad\; $ s.t.\;   $\|
    \mb{Z}_{\ell+1,t'}^j\|_F^2 = m_j$
    \EndFor

    \State \textbf{Output: } Network $\mb \Theta$ with parameters $\mb E_\ell$ and $\{\mb C_\ell^j\}$
    \end{algorithmic}\label{algo}
\end{algorithm} 

\subsection{Comparison to Existing Methods}
Incremental learning with ReduNet offers several nice properties: 
1) Each parameter of the network has an explicit purpose, computed precisely to emulate the gradient ascent on the feature representation. 
2) It does not require a memory buffer which is often needed in many state-of-the-art methods \cite{icarl,bic,benchmark}. 
3) It can be proven to behave like a network reconstructed from joint training, thus eliminating the problem of catastrophic forgetting.

Note that many existing works without relying on exemplars \cite{lwf,ewc,oewc,si} regularize the original weights of the model at each training session, effectively freezing certain parts of the network. Different tasks, however, tend to depend on different parts of the network, which eventually leads to conflicts on which parameters to regularize as the number of tasks to learn increases.
These methods, as we see later in Figure \ref{fig:accfig}, empirically perform sub-optimally in the class-IL setting. This in fact reveals the fundamental limitation that underlies in many incremental learning methods: a lack of understanding of how individual weights impact the learned representation of data points. ReduNet, on the other hand, sidesteps this problem by utilizing a fully interpretable architecture.

One notable property of ReduNet, at its current form, is that its width grows linearly with the number of classes as a new compression term $\mb C_\ell^j$ is appended to each layer whenever we see a new class. 
On the surface, this makes ReduNet similar to some architecture-based methods \cite{ltg,pnn} that dynamically expand the capacity of the network. However, there exists a major difference. ReduNet is naturally suited for the class-IL scenario, whilst the aforementioned works do not address class-IL directly. Instead they only directly address task-IL, which they accomplish by optimizing a sub-network \textit{per task}. These networks, which are designed to accomplish each task individually, fail to properly share information between the sub-networks to discriminate between classes of different tasks. ReduNet accomplishes class-IL by not only appending the class compression terms $\mb C^j$ to the network, but also modifying the expansion term $\mb E_\ell$ to share information about classes of all previously seen tasks.

For class-IL, such methods that also append new parameters to the architecture fail to completely address the problem of catastrophic forgetting. One can see why with a simple example. Consider an ensemble learning technique where for each class $j$, we train an all-versus-one model that predicts whether a data point belongs to class $j$ or not. At each task, we can feed the available data points into each model, labeled as 1 if it belongs in that class or 0 otherwise. However, by optimizing such ``black box'' models by back-propagation, we again arrive at the problem of catastrophic forgetting. Specifically, the model only sees training points of its own class only for one task or training session. For the remaining tasks, all data points that it must train will be of label 0, which prevents standard gradient descent from correctly learning the desired all-versus-one decision boundary, and there is no clear way to precisely address this optimization problem.

Although it is natural to expect the network to expand as the number of classes increases, it remains interesting to see if the growth of certain variations of the ReduNet can be sublinear instead of linear in the number of classes. 

\section{Experiments}
We evaluate the proposed method on MNIST and CIFAR-10 datasets in a class-IL scenario and compare the results with existing methods. In short, for both MNIST and CIFAR-10, the 10 classes are split into 5 incremental batches or tasks of 2 classes each. After training on each task, we evaluate the model's performance on test data from all classes the model has seen so far. The same setting is applied to all other methods we compared to. 

\subsection{Datasets} We compare the incremental learning performance of ReduNet on the following two standard datasets.

\noindent \textbf{MNIST \cite{lecun1998gradient}}. MNIST contains 70,000 greyscale images of handwritten digits $0$-$9$, where each image is of size 28 $\times$ 28. 
The dataset is split into training and testing sets, where the training set contains 6,000 images for each digit and the testing dataset contains 1,000 images for each digit. 

\noindent \textbf{CIFAR-10 \cite{krizhevsky2009learning}}. CIFAR-10 contains 60,000 RGB images of 10 object classes, where each image is of size 32 $\times$ 32. Each class has 5,000 training images and 1,000 testing images. We normalize the input data by dividing the pixel values by 255, and subtracting the mean image of the training set. 

\subsection{Implementation Details}
We implement ReduNet for each training dataset in the following manner.

\noindent \textbf{ReduNet on MNIST}. 
To construct a ReduNet on MNIST, we first flatten the input image and represent it by a vector of dimension $784$.
Then, with a precision $\epsilon$ = 0.5 in the MCR$^2$ objective \eqref{eq:mcr}, we apply 200 iterations of projected gradient iterations to compute $\mb E_\ell$ and $\mb C^j_\ell$ matrices for each iteration $\ell$.
The learning rate is set to $\eta = 0.5 \times 0.933^\ell$ at the $\ell$-th iteration. 
These matrices are the parameters of the constructed ReduNet.
Given a test data, its feature can be extracted with the incremental transform in \eqref{eq:increment} with estimated labels computed as in \eqref{eq:estimated-membership} with parameter $\lambda = 1$. At each training session, we update the ReduNet by the procedure described in Algorithm~\ref{algo}.

We note that hyper-parameter tuning in ReduNet does not require a training/validation splitting as in regular supervised learning methods. The hyper-parameters described above for ReduNet are chosen based on the training data. This is achieved by evaluating the estimated label through \eqref{eq:estimated-membership} on the training data, and comparing such labels with ground truth labels. 
Then, the model parameter $\epsilon$, learning rate $\eta$ and the softmax confidence parameter $\lambda$ are chosen as those that gives the highest accuracy with the estimated labels (at the final layer). 

\noindent \textbf{ReduNet on CIFAR-10}. 
We apply $5$ random Gaussian kernels with stride 1, size $3\times 3$  on the input RGB images.\footnote{This choice is limited by our current computational resources. Although this choice is not adequate to achieve top classification performance, it is adequate to verify the advantages of our method in the incremental setting.} 
This lifts each image to a multi-channel signal of size $32\times 32 \times 5$, which is subsequently flattened to be a $\R^{5,120}$ dimensional vector. 
Subsequently, we construct a 50-layer ReduNet with all other hyper-parameters the same as those for MNIST. 
All hyperparameters stated above, including the depth of the network, were chosen such that the $\Delta R$ loss has sufficiently converged.

\noindent \textbf{Comparing Methods.} 
We compare our approach to the following state of the art algorithms: iCaRL \cite{icarl}, LwF \cite{lwf}, oEWC \cite{oewc}, and SI \cite{si}. For these algorithms, we utilize the same benchmark and training protocol as Buzzega \etal \cite{benchmark}. For MNIST, we employ a fully-connected network with two hidden layers comprised of 100 ReLU units. For CIFAR-10, we rely on ResNet18 without pre-training \cite{resnet18}. All the networks were trained by stochastic gradient descent. For MNIST, we train on one epoch per task. For CIFAR-10, we train on 100 epochs per task. The number of epochs were chosen based on the complexity of the dataset. For each algorithm,  batch size, learning rate, and specific hyperparameters for each algorithm were selected by performing a grid-search using 10\% of the training data as a validation set and selecting the hyperparameter that achieves the highest final accuracy. The optimal hyperparameters utilized for the benchmark experiments are reported in \cite{benchmark}. 

The performance of state of the art algorithms utilizing a replay buffer highly depends on the number of exemplars, or samples from previous tasks, it is allowed to retain. We test on one exemplar-based algorithm, iCaRL. For both MNIST and CIFAR-10, we set the total number of exemplars to 200.

\subsection{Nearest Subspace Classification} \label{nearsub}
By the principle of maximal rate reduction, the ReduNet $f(\mb X,\theta)$ extracts features such that each class lies in a low-dimensional linear subspace and different subspaces are orthogonal. As suggested by the original $\text{MCR}^2$ work \cite{redunet}, we utilize a nearest subspace classifier to classify the test data featurized to maximize $\Delta R$. Given a test sample $\mb z_{test} = f(\mb x_{test},\theta)$, the label predicted by a nearest subspace classifier is

\begin{equation}
    y = \argmin_{y \in {1,...,k}}\normsq{(\mb I - \mb U^y \mb U^{y\top})\mb z_{test}}{2}{2},
\end{equation}

\noindent where $\mb U^y$ is a matrix containing the top $x$ principle components of the covariance of the training data passed through ReduNet, \ie $\mb Z_{train} \mb Z_{train}^\top$ for $\mb Z_{train} = f(\mb X_{train},\theta)$.

Since we do not have access to $\mb Z_{train}$ during evaluation, we instead collect the ${\mb C}^j$ matrices at the very last layer $L$ and extract the covariance matrix ${\mb \Sigma}_L^j$ to be further processed by SVD. For MNIST, we utilize the top $28$ principle components. For CIFAR-10, we utilize the top $15$ principle components.

\begin{table*}[!htb]
  \centering
  \caption{Incremental learning results (accuracy $\%$) on
MNIST and CIFAR-10.\vspace{0.2in}}
  \begin{tabular}{@{}c|lllll|clllll@{}}
    \hline 
     Algorithm
    &\multicolumn{5}{c}{\textbf{MNIST}}
    &\multicolumn{5}{c}{\textbf{CIFAR-10}} &\\
    \cline{2-6} \cline{7-12} 
    & Task 1 & Task 2 & Task 3 & Task 4 & Task 5 && 
    Task 1 & Task  2 & Task 3 & Task 4 & Task 5\\ 
   \hline
   LwF \cite{lwf} & 0.999&0.494 &0.333	&0.252	&0.196 & & 0.979 & 0.461 &0.319 &0.247	&0.196 \\
   oEWC \cite{oewc} & \textbf{1.0} &	0.491 &	0.332 &	0.25 &	0.217 & & 0.981&	0.383&	0.296 &	0.246 &	0.194\\
   SI \cite{si} & 0.997&	0.494&	0.411&	0.297&	0.197 && \textbf{0.989}&	0.461&	0.317&	0.248&	0.195\\
   iCaRL \cite{icarl} &0.999&0.889&0.8&0.768&0.7&&0.964&0.662&0.547&	0.5&	0.48\\
   ReduNet(Ours) & 0.999& \textbf{0.990}& \textbf{0.984} & \textbf{0.975}& \textbf{0.961}&& 0.875& \textbf{0.678}&\textbf{0.588}&\textbf{0.547}&\textbf{0.539}\\
   \hline
 \end{tabular}
 \label{fig:acctab}
\end{table*}
\begin{figure}[!htb]
\begin{center}
    \includegraphics[width=0.97\columnwidth]{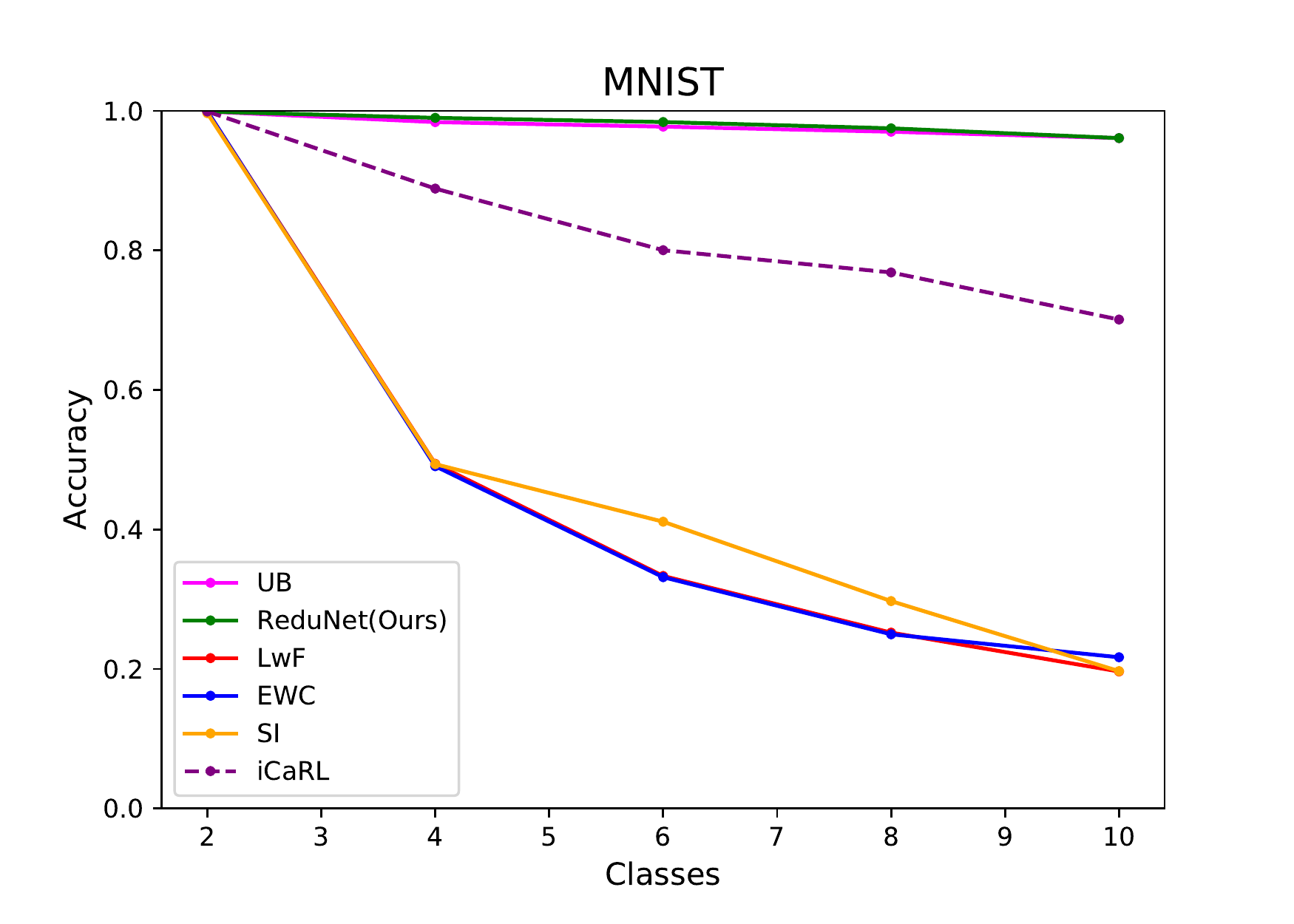}
    \includegraphics[width=0.97\columnwidth]{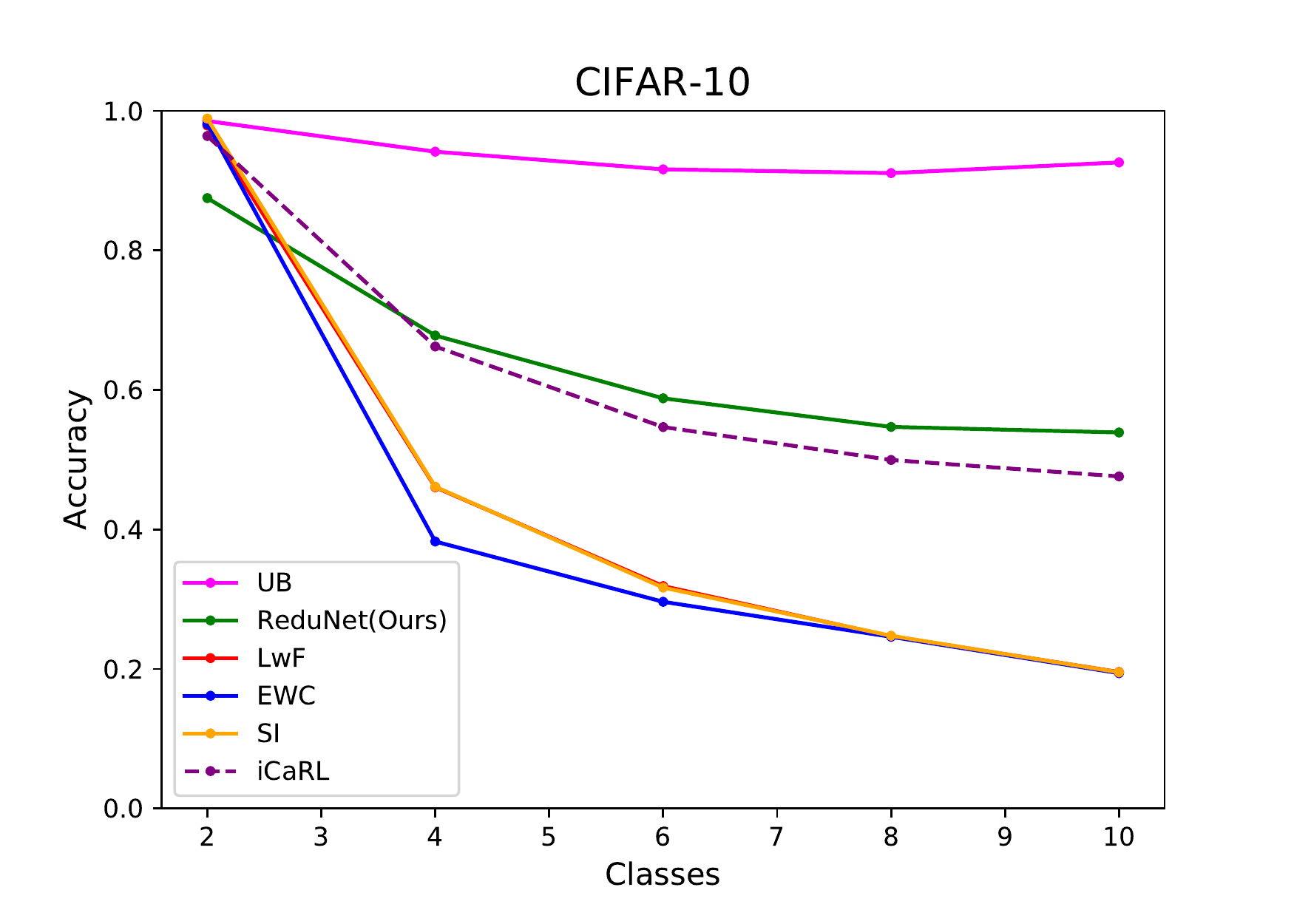}
 \end{center}
  \caption{Incremental learning results (accuracy $\%$) on
MNIST and CIFAR-10. Both datasets have 5 incremental batches. We also provide the upper bound (UB) of joint training a model utilizing the same architecture as the baseline methods. In \textit{solid} lines are regularization-based methods and in \textit{dashed} are exemplar-based methods, which saves samples from previous tasks. Note that the decay in the performance in ReduNet is simply because classification is harder to accomplish with more classes, not because of catastrophic forgetting. }
    \label{fig:accfig}
\end{figure}

\subsection{Results and Analysis}

 In this section, we evaluate the class-IL performance of incremental ReduNet against three regularization-based methods (oEWC, SI, LwF) and one method leveraging 200 exemplars (iCaRL) on MNIST and CIFAR-10. After the model trains on each task, performance is evaluated by computing the accuracy on test data from all classes the model has seen so far. For both MNIST and CIFAR-10, we observe a substantial performance increase by utilizing incremental ReduNet as shown in Figure \ref{fig:accfig} and Table \ref{fig:acctab}. 
 
 On MNIST, we observe a 3\% decay in accuracy across the tasks on ReduNet versus a 20-80\% decay in accuracy on benchmark methods (see Table \ref{fig:acctab}). We measure \textit{decay} as the difference in average accuracy between the first and last task. ReduNet retains a classification accuracy of 96\%. This is of no surprise since MNIST is relatively linearly separable, allowing second-order information about the data to be sufficient for ReduNet to correctly classify the digits. We observe that even for a very simple task as MNIST, the state of the art continual learning algorithms fail spectacularly due to catastrophic forgetting. The surprisingly large decay in the performance of benchmark methods is expected, replicated often in class continual learning literature \cite{benchmark, yu2020drift}. Note that ReduNet observes no catastrophic forgetting and the decay in its performance is simply because classification is increasingly harder to accomplish with more classes.

Surprisingly, we still observe an improvement in performance utilizing ReduNet on CIFAR-10, a more complex image dataset. We observe a 45-80\% decay in accuracy for benchmark methods, whereas incremental ReduNet observes a 34\% decrease (in Table \ref{fig:acctab}). The algorithm that achieves the closest performance to ReduNet is iCaRL, an exemplar-based method that requires access to 200 exemplars it has previously observed. Certainly, as can be seen by the 88\% accuracy on Task 1 of CIFAR-10, ReduNet at its current basic form (only using 5 randomly initialized kernels, no back-propagation training) is not able to reach the same classification accuracy as ResNet-18 for complex image classification tasks. However, ReduNet decays gracefully and significantly outperforms other deep learning methods in the continual learning setting, as other methods suffer from catastrophic forgetting as it acquires knowledge from new tasks.

\section{Conclusions and Future Work}
In this work, we have demonstrated through an incremental version of the recently proposed ReduNet, the promise of leveraging interpretable network design for continual/online learning. The proposed network has shown significant performance increases in both synthetic and complex real data, even without utilizing any fine-tuning with back-propagation. It has clearly shown that if knowledge of past learned tasks are properly utilized, catastrophic forgetting needs not to happen as new tasks continue to be learned. 

We want to emphasize that it is not the purpose of this work to push the state of art classification accuracy or efficiency on any single large-scale real-world task or dataset. Rather we want to use the simplest experiments to show beyond doubt the remarkable effectiveness and great potential of this new framework. Using CIFAR-10 as an example, simply utilizing a relatively small set of 5 random lifting kernels was already sufficient for a decent incremental classification performance. We believe that to achieve better performance or for more complex tasks and datasets, judicious design or learning of more convolution kernels would be needed. This leaves plenty of room for further improvements.

This work also opens up a few promising new extensions. As we have mentioned earlier, the current framework requires the width of the network to grow linearly in the number of classes. It would be interesting to see if some of the filters can be shared among old/new classes so that the growth can be sublinear. To a large extent, the rate reduction gives a unified measure for learning  discriminative representations in supervised, semi-supervised, and unsupervised settings. We believe our method can be easily extended to cases when some of the new data do not have class information. 

\newpage
{\small
\bibliographystyle{ieee_fullname}
\bibliography{egbib}
}

\end{document}